\documentclass{bmvc2k}
\usepackage{graphicx}
\usepackage{amsthm}
\usepackage{amsfonts}
\usepackage{algorithm}
\usepackage{algpseudocode}
\usepackage{hyperref}

\title{Equivariance-bridged SO(2)-Invariant Representation Learning using \\ Graph Convolutional Network}

\addauthor{Sungwon Hwang}{shwang.14@kaist.ac.kr}{1}
\addauthor{Hyungtae Lim}{shapelim@kaist.ac.kr}{1}
\addauthor{Hyun Myung}{hmyung@kaist.ac.kr}{1}

\addinstitution{
 School of Electrical Engineering, \\ Korea Advanced Institute of \\ Science and Technology (KAIST), \\
 Daejeon, Korea
}

\runninghead{Hwang, Lim, Myung}{SO(2)-Invariant Representation Learning}

\begin{document}

\maketitle

\begin{abstract}
Training a Convolutional Neural Network (CNN) to be robust against rotation has mostly been done with data augmentation. In this paper, another progressive vision of research direction is highlighted to encourage less dependence on data augmentation by achieving structural rotational invariance of a network. The deep equivariance-bridged SO(2) invariant network is proposed to echo such vision. First, Self-Weighted Nearest Neighbors Graph Convolutional Network (SWN-GCN) is proposed to implement Graph Convolutional Network (GCN) on the graph representation of an image to acquire rotationally equivariant representation, as GCN is more suitable for constructing deeper network than spectral graph convolution-based approaches. Then, invariant representation is eventually obtained with Global Average Pooling (GAP), a permutation-invariant operation suitable for aggregating high-dimensional representations, over the equivariant set of vertices retrieved from SWN-GCN. Our method achieves the state-of-the-art image classification performance on rotated MNIST and CIFAR-10 images, where the models are trained with a non-augmented dataset only. Quantitative validations over invariance of the representations also demonstrate strong invariance of deep representations of SWN-GCN over rotations.

\end{abstract}

\vspace{-0.2cm}
\section{Introduction}
\vspace{-0.3cm}
In the past few years, Convolutional Neural Network (CNN) has brought many advances, especially on many computer vision tasks. The high performance leverages the use of learned convolution filters, with phenomenal techniques to make the layers go deeper \cite{he2016deep}. Such advancements led to the performance close to human in image classification tasks on various datasets \cite{lecun1998gradient, zeiler2014visualizing, krizhevsky2017imagenet, szegedy2015going, he2016deep}. Especially, deeper layers of CNN have been empirically shown to learn substantially more translation-invariant features in each layer, which takes account for a wide range of applicability and reliability of CNN \cite{goodfellow2009measuring, schmidt2012learning, cohen2016group}.

However, achieving rotation invariance is another desirable property of a network, especially on applications that require inferences over arbitrarily rotated images, such as aerial \cite{wang2020invariant} or biomedical microscopy images \cite{weiler2018learning}. To do so, the most common practice to train a neural network to yield rotation-invariant representations is data augmentation \cite{perez2017effectiveness}. By providing rotation-augmented images for training, a network can learn representations expressed in different rotations. If associated with a correct objective function, the network can yield reasonably invariant representations or inferences regardless of the rotation of an input \cite{shorten2019survey}. 

Yet, models trained with data augmentation may fail to capture local equivariance and entail the black-box problem \cite{worrall2017harmonic}. Besides, with an extensive list of data augmentations to train a transformation-robust network, the exponential growth of search space for augmentations inflates the dataset size \cite{shorten2019survey}. Thus, liberation from rotation augmentation during training not only decreases a substantial number of training but can also give more search space for other types of data augmentations. Considering that rotation is one of the most common types of data augmentations \cite{shorten2019survey}, achieving rotation-invariance of a network is a large leap towards less dependence on data augmentation. 

To address this issue, TIGraNet \cite{khasanova2017graph} made the most recent attempt to explicitly define the aforementioned problem and to suggest a solution. The researchers validated the transformation invariance of their proposed network by training their network only with upright 2D images with no transformation augmentation and evaluating the network with images augmented with isometric transformations. We extend their view and propose an alternative over their spectral graph convolution-based method to construct deeper image representation network using an equivariance-bridged SO(2) invariant graph convolutional network.


    

\vspace{-0.4cm}
\section{Related Works}
\vspace{-0.3cm}

\paragraph{Transformation-Equivariant Networks} Instead of obtaining multiple filters that represent different rotations, steerable filters can be constructed with finite linear combinations of irreducible representations in order to achieve transformation equivariance \cite{freeman1991design}. The recent work extends the concept to construct steerable CNN by obtaining homomorphisms of transformations built with base representations from transformation-equivariant filter bank \cite{cohen2016steerable}. 
Following works include parametrizing the steerable filters \cite{weiler2018learning}, restricting the filters to be of the form from the circular harmonic family to achieve hard-baked rotation equivariance \cite{worrall2017harmonic}, or applying convolutional filters at multiple orientations to retrieve vector field representations of deep features \cite{marcos2017rotation}.

The most recently, E(2)-CNN \cite{weiler2019general} makes a holistic implementation of the aforementioned transformation-equivariant networks on \textcolor{black}{the} steerable filters. Based on the group theory, the implementation achieves E(2) (translation, rotation and reflection in Euclidean space) equivariance and achieves the state-of-the-art performance over MNIST rot \cite{larochelle2007empirical} dataset classification task.

\vspace{-0.4cm}
\paragraph{Deep Learning on Graphs} Promising potentials of graph-based networks were demonstrated from the generalization of CNN to low-dimensional graph domains along with the extension of the convolution via the Laplacian Spectrum \cite{bruna2013spectral}. Then, spectral network with a graph estimation procedure enabled the graph-based network to go deeper, demonstrating its superior performance over large-scale classification problems \cite{henaff2015deep}.

Meanwhile, GCN \cite{kipf2016semi} has been one of the most prevalent graph processing networks and was devised as renormalized first-order approximations of spectral graph convolutions to conduct semi-supervised learning on graph-structured data. 
GCN has been effectively applied to a wide range of fields, such as but not limited to multi-label image recognition \cite{chen2019multi}, temporal action localization \cite{zeng2019graph}, and even solid-state material science \cite{schmidt2019recent}. 

\vspace{-0.4cm}
\paragraph{Transformation-Invariant Networks} Ti-Pooling \cite{laptev2016ti} achieves rotation invariant representation by feeding multiple rotated instances of an image to a siamese network, followed by element-wise pooling. Meanwhile, Spatial Transformer Network (STN) achieves the capacity to yield transformation-invariant representation by learning the affine transformation within the data with much distortion \cite{jaderberg2015spatial}. In the meantime, graph-based isometry-invariant network \cite{khasanova2017graph} was proposed as a successful attempt to represent an image to be isometry-invariant using graph. Their methodology uses spectral convolutions and dynamic pooling to retrieve isometry-equivariant graph representation of an image, followed by a statistical layer over Chebyshev polynomial representations of graph signals to retrieve the isometry-invariant representations. 


\vspace{-0.2cm}
\section{Problem Definition}

\vspace{-0.3cm}
\subsection{Equivariance and Invariance} \label{subsec:equiinv}
\vspace{-0.2cm}

Given a function $f:X \rightarrow Y$, $f$ is said to be \textit{equivariant} to a group of transformations if every transformation $\psi \in \Psi$ of an input $\mathcal{X} \in X$ can be associated with an equal transformation 
of the corresponding representation \textcolor{black}{$\mathcal{Y} \in Y$}, or

\vspace{-0.3cm}

\begin{equation}
    f(\psi(\mathcal{X})) = \psi(f(\mathcal{X})) = \psi(\mathcal{Y}).
\end{equation}

Meanwhile, given a function $g:  Y \rightarrow Z$ 
, $g$ is said to be \textit{invariant} over transformation $\psi$ if $\psi$ in space $Y$ yields identity transformation in $Z$, or

\vspace{-0.3cm}

\begin{equation}
    g(\psi(\mathcal{Y}))=g(\mathcal{Y}).
\end{equation}

Then, the composition of equivariant function $f$ followed by invariant function $g$ eventually yields an invariant function with respect to $\psi$ as follows:

\vspace{-0.3cm}
\begin{equation} \label{eq:equiinv}
    g(f(\psi(\mathcal{X})))=g(\psi(f(\mathcal{X})))=g(f(\mathcal{X})).
\end{equation}

We bridge the equivariant network as the means to effectively project an input to a high-dimensional representation space instead of directly transforming an input image to invariant representation. The illustrative explanation can be found in Figure \ref{fig:def}-(a).

\vspace{-0.2cm}
\subsection{Rethinking the Value of Equivariant Networks}
\vspace{-0.2cm}

Most methodologies on equivariant networks \cite{weiler2019general, worrall2017harmonic} validate their performances by training and testing their networks over rotation-augmented datasets, such as MNIST-rot \cite{larochelle2007empirical}. The significance of the works is that the equivariant networks have larger capacity to learn all the different representations of rotation augmented inputs, since the representations are less variational and more equivariant, whose consistencies are easier to be adapted.

However, our objective lies on constructing a structurally invariant network that can make invariant inferences over rotations without rotation augmentations. Acquiring high-dimensional and equivariant feature space is a key step to achieving this goal by associating with a transformation-invariant function, as will be described in Section~\ref{subsec:objective}.

\vspace{-0.1cm}
\subsection{Equivariance-bridged SO(2) Invariant Network}
\vspace{-0.2cm}
\label{subsec:objective}
Specifying $X$ as a training dataset and given an objective function $\mathcal{L}$, our goal is to find a rotation-equivariant network $f(\cdot ; w_{f})$, rotation-invariant function $g(\cdot)$, and optimal parameters $w^{*}_{f}$ of the network $f$ that satisfy

\vspace{-0.3cm}
\begin{equation}
w_{f}^{*} = \underset{w_{f}}{\mathrm{argmin}} \; \mathcal{L}(g(f(X;w_{f}))).
\end{equation}

Then, given a rotation angle $\theta \in [0^{\circ}, 360^{\circ})$, a corresponding rotation transformation $\textbf{R}^{\theta} \in \textbf{R}$, where $\textbf{R}$ is a group of rotation transformation on image representation and forms a homomorphism with SO(2), and a correspondingly rotated image $\mathcal{R}^{\theta}=\textbf{R}^{\theta}(\mathcal{X})$, our objective network and parameters should satisfy from Eq. \eqref{eq:equiinv} as

\vspace{-0.3cm}
\begin{equation}\label{eq:obj_2}
    g(f(\mathcal{R}^{\theta};w^{*}_{f})) = g(f(\mathcal{X};w^{*}_{f})).
\end{equation}

\begin{figure}
\begin{tabular}{cc}
\bmvaHangBox{\fbox{\includegraphics[width=5.6cm]{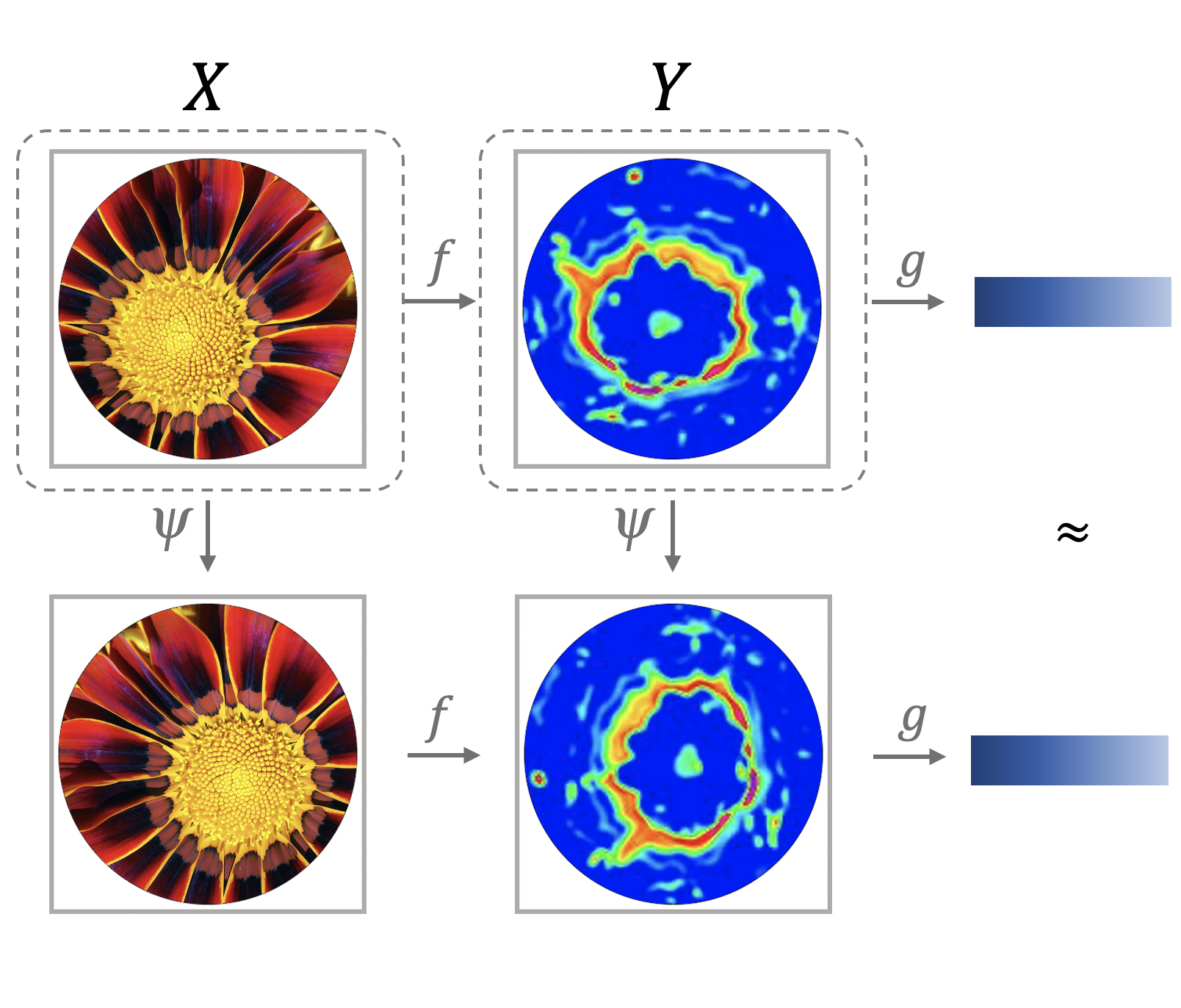}}}&
\bmvaHangBox{\fbox{\includegraphics[width=5.8cm]{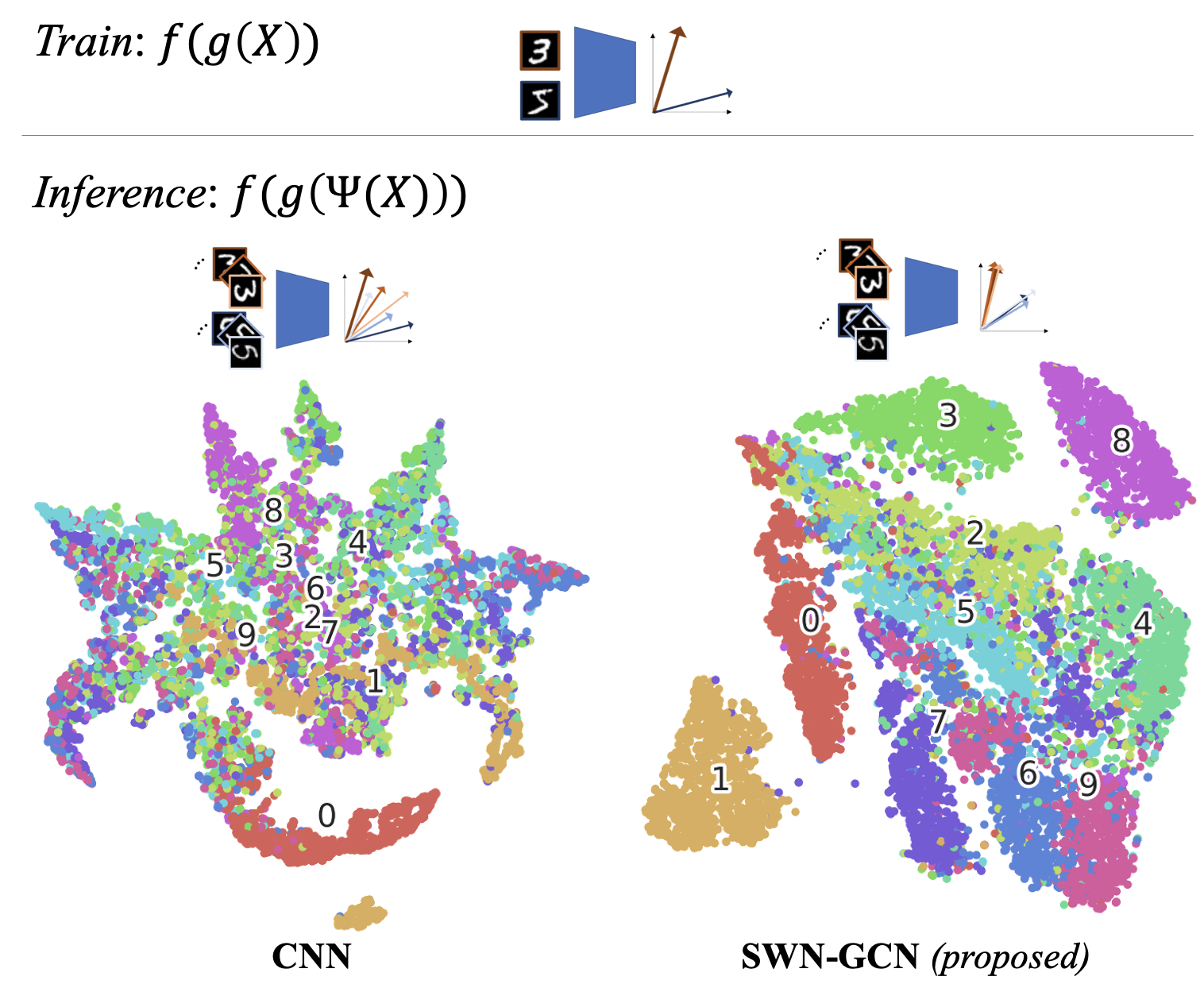}}}\\
(a)&(b)
\end{tabular}
\caption{(a) Illustration of equivariant function $f$, invariant function $g$, transformation function $\psi$, and their compositional relationships. Composition of $f$ and $g$ yields a $\Psi$-invariant representation (blue representation boxes); (b) Illustration of our objective and comparison of t-SNE \cite{maaten2008visualizing} visualized feature representations of randomly rotated MNIST extracted with Convolutional Neural Networks (CNN) (left) and our proposed method (right). Both models are trained with non-augmented, upright MNIST dataset.}
\label{fig:def}
\end{figure}

Thus, $g(f(\cdot;w^{*}_{f}))$ trained with \textcolor{black}{the} upright dataset $X$ should be able to yield invariant representation for every $\mathcal{R}^{\theta}$. Such objective and t-SNE \cite{maaten2008visualizing} visualized representations of randomly rotated input images extracted from our method compared with those extracted from ResNet-50 are available in Figure \ref{fig:def}-(b).

\section{Proposed Method}
\vspace{-0.3cm}

\begin{figure*}
  \includegraphics[width=0.9\linewidth]{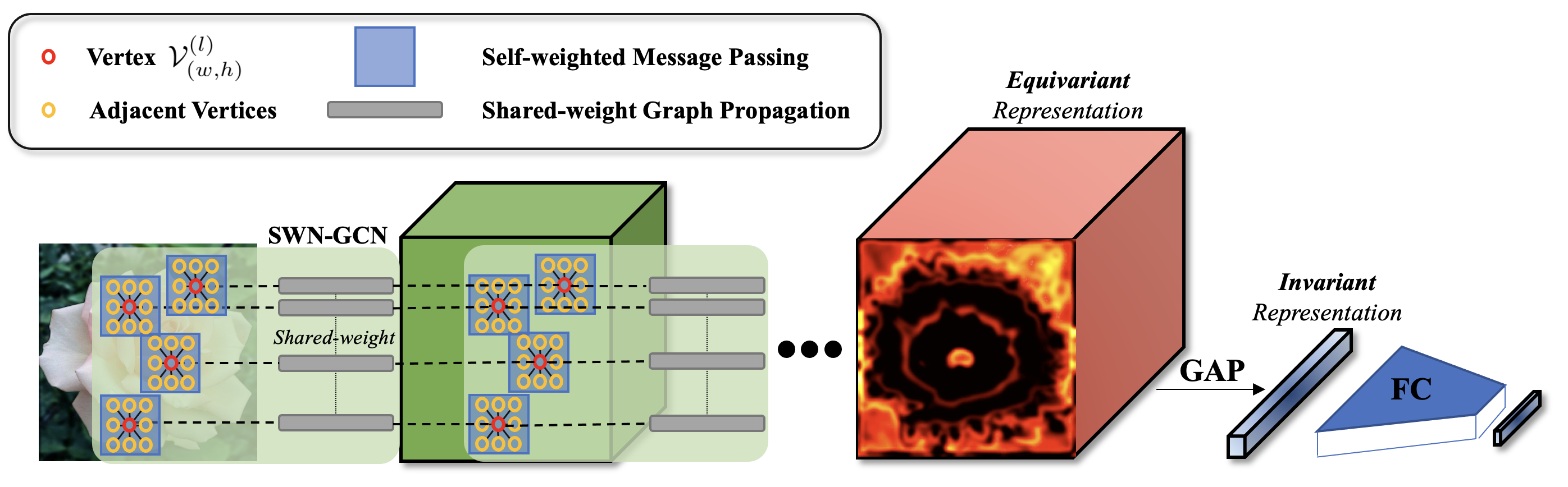}
  \centering
  \vspace{-0.4cm}
  \caption{Self-Weighted Nearest-Neighbors Graph Convolutional Network (SWN-GCN).}
  \label{fig:overview}
\end{figure*}

The schematics of our proposed network is summarized in Figure \ref{fig:overview}. Self-Weighted Nearest-Neighbors Graph Convolutional Network (SWN-GCN) is proposed to learn graph-based features in high dimensional and rotation-equivariant representation space, followed by Global Average Pooling (GAP) for invariance mapping of equivariant representations.


\vspace{-0.1cm}
\subsection{SWN-GCN}
\label{section:SWN-GCN}
\vspace{-0.2cm}

\paragraph{Propagation Rule}  
GCN \cite{kipf2016semi} is applied over an instance of image $\mathcal{X}$ of width $W$ and height $H$ expressed as an undirected graph representation $\mathcal{G} = (\mathcal{V}, A)$. $\mathcal{V}$ is a set of $|\mathcal{V}|=W \cdot H$ number of vertices and $A \in \mathbb{R}^{|\mathcal{V}| \times |\mathcal{V}|}$ is the adjacency matrix between the vertices.
Specifically, we start off by representing $\mathcal{X}$ as a graph with vertices of 

\vspace{-0.2cm}
\begin{equation}\label{eq:vertex}
   \mathcal{V}^{(0)}=[\mathcal{V}_{(1, 1)}^{(0)} \cdots \mathcal{V}_{(w, h)}^{(0)} \cdots \mathcal{V}_{(W, H)}^{(0)}]^{\top} \in \mathbb{R}^{|\mathcal{V}| \times c^{(0)}}
\end{equation}

where $\mathcal{V}_{(w, h)}^{(0)}$ denotes \textcolor{black}{channel-wise values} located at $(w, h)$ in image $\mathcal{X}$ and $c^{(0)}$ is the size of input channel of $\mathcal{X}$, i.e. $c^{(0)}=1$ for gray-scale images and $c^{(0)}=3$ for color images. The component of the adjacency matrix $A$ that represents \textcolor{black}{the adjacency} between two vertices $\mathcal{V}^{(0)}_{(w, h)}$ and $\mathcal{V}^{(0)}_{(w^{'}, h^{'})}$ is denoted as ${A}_{[(w,h),(w^{'}, h^{'})]}$ and is defined as follows:

\vspace{-0.2cm}
\begin{equation}\label{eq:adj}
    {A}_{[(w, h),(w', h')]} =  
    \begin{cases}
    1 & \text{if} \ 0 < d_{(w, h)}^{(w', h')} \leq \sqrt{2} \\
    0 & \text{otherwise}
    \end{cases}
\end{equation}

\textcolor{black}{where $d_{(w, h)}^{(w', h')} = \sqrt{(w - w')^{2}+(h - h')^{2}}$. 
}

Then, given \textcolor{black}{the} $l$-th propagated set of vertices $\mathcal{V}^{(l)}$, a unit propagation of SWN-GCN comprises the two networks, which we will refer to as \textit{Self-weighted Message Passing} network (SMP, $\zeta(\cdot)$) and \textit{Shared-weight Graph Propagation} network (SGP, $\xi(\cdot)$), to construct a unit layer of our equivariant network $f^{(l)}: \mathcal{V}^{(l)} \rightarrow \mathcal{V}^{(l+1)}$ correspondingly as

\vspace{-0.2cm}

\begin{equation}
    \hat{\mathcal{V}}^{(l)} = \bar{D}^{-\frac{1}{2}} \bar{A}^{(l)} \bar{D}^{-\frac{1}{2}}\mathcal{V}^{(l)} := \zeta^{(l)}(\mathcal{V}^{(l)}, A) \label{eq:hatprop}
\end{equation}


\vspace{-0.2cm}
\begin{equation}
    \textcolor{black}{\mathcal{V}^{(l+1)}=\kappa^{(l)}_{2}(\kappa^{(l)}_{1}(\hat{\mathcal{V}}^{(l)})) := \xi^{(l)}(\hat{\mathcal{V}}^{(l)})} \label{eq:nextprop}
\end{equation}

where $\bar{A}^{(l)} = A + \beta^{(l)} I_{|\mathcal{V}|}$ is the self-weighted adjacency matrix and $\beta^{(l)}$ is a trainable parameter. $\bar{D} \in \mathbb{R}^{|\mathcal{V}| \times |\mathcal{V}|}$ is a diagonal matrix formulated as:


\vspace{-0.2cm}
\begin{equation}
    \bar{D}_{[(w,h),(w,h)]}=
    1 + \sum_{w', h'=1}^{w' = W, h' = H} A_{[(w,h), (w', h')]}.\label{eq:degreedef}
\end{equation}

The trainable function $\kappa^{(l)}_{i}(\cdot)$ in SGP is defined as

\vspace{-0.3cm}
\begin{equation}
    \textcolor{black}{\kappa^{(l)}_{i}(\cdot)=\sigma(BN^{(l)}_{i}(\; \cdot \; W^{(l)}_{i}))}
\end{equation}

\textcolor{black}{where $c'^{(l)}, W_{1}^{(l)}\in \mathbb{R}^{c^{(l)} \times c'^{(l)}}$, $ W_{2}^{(l)} \in \mathbb{R}^{c'^{(l)} \times c^{(l+1)}}$, $BN_{i}$, and $\sigma$ are intermediate dimension size, first and second propagation parameters, batch-normalization \cite{ioffe2015batch}, and ReLU non-linearity, respectively.}

\vspace{-0.3cm}

\paragraph{Leveraging GCN to construct deeper model} Given a diagonal matrix $D$ where $D_{ii} = \sum_{j}A_{ij}$, if the linear approximation of Chebyshev polynomials of spectral graph convolution is applied over our method, SMP would be formulated as $\hat{\mathcal{V}}^{(l)} = (I_{|\mathcal{V}|} + D^{-\frac{1}{2}}AD^{-\frac{1}{2}})\mathcal{V}^{(l)}$. However, eigenvalues of $I_{|\mathcal{V}|} + D^{-\frac{1}{2}}AD^{-\frac{1}{2}}$ range in $[0, 2]$, which means that multiple stacks of these layers to construct deeper models may cause exploding or vanishing gradient problems. GCN tackled this concern directly and conducted renormalization of the operation by substituting $I_{|\mathcal{V}|} + D^{-\frac{1}{2}}AD^{-\frac{1}{2}}$ with $\tilde{D}^{-\frac{1}{2}}\tilde{A}\tilde{D}^{-\frac{1}{2}}$, where $\tilde{A} = A + I_{|\mathcal{V}|}$ and $\tilde{D}_{ii} = \sum_{j}\tilde{A}_{ij}$, thus allowing the network to back-propagate more stable in deeper layers. We additionally included batch normalization to reduce the typical covariance shift problem \cite{ioffe2015batch} in deeper models. More details on the derivation of the approximation of spectral graph convolution and the following discussion can be found in \cite{kipf2016semi}.

\vspace{-0.1cm}
\subsection{SO(2)-Equivariant Property of SWN-GCN} 
\vspace{-0.2cm}

In this subsection, we show that the architecture of SWN-GCN yields rotation-equivariant representation by its structural nature. Primarily, representation of a rotated image must be strictly defined.

\textbf{Definition 1.} \textit{Given} $(u', v')$ \textit{and} $\textbf{T}^{\theta} \in SO(2)$ \textit{that satisfies} $\textbf{T}^{\theta}[w-\frac{W}{2}, h-\frac{H}{2}]^T=[u'-\frac{W}{2}, v'-\frac{H}{2}]^T$, \textit{every pixel value of rotated image} $\mathcal{R}^{\theta}_{(u, v)}$ \textit{is defined as}:

\vspace{-0.2cm}
\begin{equation}
     \mathcal{R}^{\theta}_{(u, v)} := h(\mathcal{X}_{(w, h)} \label{eq:rotdef} ; (u', v'))
\end{equation}

\textit{where} $h(\cdot ; (u', v'))$ \textit{is an interpolation function}  \cite{thevenaz2000image} \textit{to assign interpolated values to the closest pixel at} $(u, v)$ \textit{in} $\mathcal{R}^{\theta}$. \textit{We may say that $\mathcal{R}^{\theta}_{(u, v)}$ and $\mathcal{X}_{(w, h)}$ forms a spatial correspondence}.

Then, we can straightforwardly show rotation equivariance of the $l$-th propagated vertex set $\mathcal{V}^{(l)}$ by showing that every vertex of an upright image and the spatially corresponding vertex of a rotated image are equal throughout the propagation as the following proposition shows.  

\textbf{Proposition 1.} \textit{Let} $\mathcal{H}$ \textit{be the vertex set representation of a rotated image} $\mathcal{R}^{\theta}$. \textit{Then, for all } $w \in \{ 1 \cdots W\}$, $h \in \{ 1 \cdots H\}$ \textit{and} $l \in \{ 0 \cdots L_{f} \}$, \textit{the following approximation holds}
\begin{equation} \label{eq:equiproof}
    \mathcal{H}^{(l)}_{(u, v)}  \approx \mathcal{V}_{(w, h)}^{(l)}.
\end{equation}

\begin{proof}
Inductive method is used to prove the proposition.


(a) When $l = 0$:  or when every vertex representation is the original pixel value, we can safely make the following assumption:

\textbf{Assumption 1.} $\mathcal{H}^{(0)}_{(u, v)} \approx \mathcal{V}_{(w, h)}^{(0)}$ \textit{for all}  $w \in \{ 1 \cdots W\}$, $h \in \{ 1 \cdots H\}$.

The assumption is reasonable from our definition of spatial correspondence in Eq. \eqref{eq:rotdef} given that interpolation does not significantly change the value of most pixels.

(b) Then, provided that Eq. \eqref{eq:equiproof} holds when $l = n$, we need to show that the equation holds when $l = n+1$, or 

\vspace{-0.3cm}
\begin{equation}
    \xi^{(n)}(\zeta^{(n)}(\mathcal{V}^{(n)}))_{(w, h)} = \xi^{(n)}(\zeta^{(n)}(\mathcal{H}^{(n)}))_{(u, v)}.
\end{equation}

First, SMP does preserve rotational invariance of spatial correspondence, or $\zeta^{(n)}(\mathcal{V}^{(n)})_{(w, h)}$ $= \zeta^{(n)}(\mathcal{H}^{(n)})_{(u, v)}$. Since there are only 9 vertices including itself (assume that image is rotated with zero padding) that yields non-zero adjacency for every vertex according to Eq. \eqref{eq:adj}, degree matrix $\bar{D}$ in Eq. \eqref{eq:degreedef} can be approximated as $9I_{|\mathcal{V}|}$. The degree matrix can thus be approximated as a scalar multiplication of an identity matrix, from which Eq. \eqref{eq:hatprop} can be rearranged as

\vspace{-0.3cm}
\begin{equation}
     \bar{D}^{-\frac{1}{2}} \bar{A}^{(l)} \bar{D}^{-\frac{1}{2}} = \bar{D}^{-1}\bar{A}^{(l)} . \label{eq:simple}
\end{equation}

Then, Eq.~\eqref{eq:simple} can be utilized
to express $\hat{\mathcal{V}}^{(n)}_{(w, h)} = \zeta^{(n)}(\mathcal{V}^{(n)})_{(w, h)}$ as

\vspace{-0.2cm}

\begin{equation}
    \hat{\mathcal{V}}^{(n)}_{(w, h)} =  
    \frac{1}{9}(\beta^{(n)} \mathcal{V}^{(n)}_{(w, h)} + \sum_{\substack{i=w-1, j=h-1 \\ i \neq w, j \neq h}}^{w+1, h+1}\mathcal{V}^{(n)}_{(i, j)}). \label{eq:approx1}
\end{equation}

Likewise, $\hat{\mathcal{H}}^{(n)}_{(u, v)} = \zeta^{(n)}(\mathcal{H}^{(n)})_{(u, v)}$ can also be expressed with the same process we leveraged to obtain Eq. \eqref{eq:approx1} as

\vspace{-0.3cm}

\begin{equation}
    \hat{\mathcal{H}}^{(n)}_{(u, v)} = \frac{1}{9}(\beta^{(n)} \mathcal{H}^{(n)}_{(u, v)} + \sum_{\substack{i=u-1, j=v-1 \\ i \neq u, j \neq v}}^{u+1, v+1}\mathcal{H}^{(n)}_{(i, j)}). \label{eq:approx2}
\end{equation}

In fact, the expectation of sum of adjacent vertices of $\mathcal{V}_{(w, h)}^{(n)}$ and $\mathcal{H}^{(n)}_{(u, v)}$ in Eq. \eqref{eq:approx1} and Eq. \eqref{eq:approx2} are equal. 
Specifically, we have local rotational consistency as shown in the following assumption:

\textbf{Assumption 2.} \textit{(Local Rotational Consistency)}
\begin{equation}\label{eq:appeq}
    \mathbb{E}\left[ \sum_{\substack{i=w-1, j=h-1 \\ i \neq w, j \neq h}}^{w+1, h+1}\mathcal{V}^{(n)}_{(i, j)} \right] =
    \mathbb{E}\left[ \sum_{\substack{i=u-1, j=v-1 \\ i \neq u, j \neq v}}^{u+1, v+1}\mathcal{H}^{(n)}_{(i, j)} \right].
\end{equation}

Rigid rotation of an image does not change the list of adjacent vertices, which means that their sum remains constant under rigid rotation. Yet, the values of vertices retrieved from images with rotations that are not multiples of $90^{\circ}$ may be slightly different due to interpolations. However, it is reasonable to assume that the sum of the adjacent vertices, each of which has slightly deviating values from interpolation, is acceptably constant. Experimental results in the later section show that this assumption is reasonable enough to yield the most invariant representation out of all baselines \textcolor{black}{(see Section \ref{sec:inv})}. Meanwhile, we are given with $\beta^{(n)}\mathcal{V}_{(w, h)}^{(n)} \approx \beta^{(n)}\mathcal{H}^{(n)}_{(u, v)}$ from the inductive assumption, from which we can conclude $\zeta^{(n)}(\mathcal{V}^{(n)})_{(w, h)} \approx \zeta^{(n)}(\mathcal{H}^{(n)})_{(u, v)}$.


Then, $\xi^{(n)}(\mathcal{\hat{V}}^{(n)})_{(w, h)} \approx \xi^{(n)}(\mathcal{\hat{H}}^{(n)})_{(u, v)}$ straightforwardly holds because multiplied weights are shared and ReLU is not a one-to-many function. Meanwhile, batch-normalization does not strictly but acceptably preserves the approximate equality (see Section A in supplementary material for details). Thus, we can conclude  $\mathcal{V}_{(w, h)}^{(n+1)} \approx \mathcal{H}^{(n+1)}_{(u, v)}$ given $\mathcal{V}_{(w, h)}^{(n)} \approx \mathcal{H}^{(n)}_{(u, v)}$ and finalize the inductive proof.
\end{proof}

\vspace{-0.6cm}
\subsection{Global Average Pooling for Invariant Mapping and Classification of Invariant Representations}
\vspace{-0.2cm}
\label{section:gap}

Recent graph-based networks \cite{khasanova2017graph, yang2020rotation} employ the statistical layer, which computes means and variances of graph signals using graph Chebyshev polynomials of order up to $K_{\max}$, to map an equivariant vertex set to an invariant representation. Given set of vertices after the $L_{f}$-th propagation, where every vertex is represented in $\mathbb{R}^{c^{(L_f)}}$, the statistical layer calculates  
$c^{(L_f)} \cdot (K_{\max} + 1)$ number of means and variances. However, such process may be burdensome on equivariant representations expressed in high representation space retrieved from deep networks like ours, where the dimension of representation space can go up to $c^{(L_{f})} = 512$.

Meanwhile, deep network such as PointNet \cite{qi2017pointnet} or Residual Network \cite{he2016deep} demonstrated that a global permutation-invariant operation, such as max or average operation, is capable of efficiently aggregating high-dimensional representations. To this end, we employ GAP to aggregate rotation-equivariant set of vertices, $\mathcal{V}^{(L_{f})}$, to invariant representation \textcolor{black}{$\textbf{z} \in \mathbb{R}^{c^{(L_{f})}}$} as

\vspace{-0.3cm}
\begin{equation}\label{eq:gap}
    \textbf{z} = \frac{1}{W \cdot H}\sum_{w=1, h=1}^{W, H}\mathcal{V}^{(L_{f})}_{(w,h)}.
\end{equation}

Since sum is a permutation-invariant operation, GAP yields rotation-invariant $\textbf{z}$ out of rotation-equivariant vertices $\mathcal{V}^{(L_{f})}$, where the permutation occurs within the rotation-equivariant set of vertices. Unlike direct use of GAP \cite{lin2013network} that requires the last feature representation dimension equal to the number of classification classes, our equivariant space is mapped into much higher dimension before GAP and the final classification is conducted with fully connected layers non-linearized with ReLU. The summary of the methods can be found in Algorithm \ref{alg:methodology}.




\begin{algorithm}
\caption{SWN-GCN for SO(2)-invariant representation of an image}\label{alg:methodology}
\begin{algorithmic}
\State \textbf{given }  $\mathcal{X}$ (input image), $L$ (Layer configurations) 
\State $\mathcal{G} = (\mathcal{V}, A) \gets$ \textit{Image2Graph(}$\mathcal{X}$\textit{)} \Comment{Eq. \ref{eq:vertex}, Eq. \ref{eq:adj}}
\For{$l$ in $L$}
    \State $\hat{\mathcal{V}} \gets  \zeta^{(l)}(\mathcal{V}, A)$ \Comment{Eq. \ref{eq:hatprop}}
    \State $\mathcal{V} \gets  \xi^{(l)}(\hat{\mathcal{V}})$ \Comment{Eq. \ref{eq:nextprop}}
\EndFor
\State $z \gets$ \textit{GAP(}$\mathcal{V}$\textit{)}\Comment{Eq. \ref{eq:gap}}
\State \textbf{return} \textit{linear\_classifier(}$z$\textit{)}
\end{algorithmic}
\end{algorithm}

\vspace{-0.2cm}
\section{Experiment}

\vspace{-0.3cm}
\subsection{Dataset}
\vspace{-0.2cm}
\paragraph{MNIST} The MNIST dataset \cite{lecun1998gradient} comprises of images with grayscale, handwritten digits of 10 classes ranging from 0 to 9. The dataset includes 60,000 train and 10,000 test images. 
\vspace{-0.3cm}
\paragraph{CIFAR-10} The CIFAR-10 dataset \cite{Krizhevsky09learningmultiple} consists of 60,000 $32 \times 32$ images in 10 classes with 6,000 images per class. The dataset includes 50,000 training images and 10,000 test images. All images are circularly masked to yield minimum interference caused by black spaces on edges created after rotation for the sake of the experiment.

\vspace{-0.1cm}
\subsection{Experimental Setup} 
\vspace{-0.2cm}
We compare our model with the state-of-the-art E(2)-equivariant network, E(2)-CNN \cite{weiler2019general}, as well as Harmonic Network \cite{worrall2017harmonic} and TIGraNet \cite{khasanova2017graph}. The most prevalent CNNs such as VGG-19 \cite{simonyan2014very} and ResNet-50 \cite{he2016deep} are also compared. 
For E(2)-CNN, we selected $C_{8}$, as it showed the best performance out of the rotation groups from our experiment.
GAP is applied for rotation equivariant models with no specific implementation of invariance mapping. The number of parameters for all models are kept approximately constant, except for VGG-19 and ResNet-50 since these are not designed for rotation equivariance but are included as baseline models to demonstrate the degree of rotation-invariance of other state-of-the-art models. 

\vspace{-0.4cm}
\subsection{Training Details}
\vspace{-0.2cm}
$80\%$ of the training sets are used for training, and the remainders for validation dataset. Models are optimized with ADAM optimizer \cite{kingma2014adam} with learning rate of $10^{-4}$ and batch size of $64$ until the models show no improvement over validation dataset for 40 consecutive epochs. All models are trained three times, and all the reports on classification accuracies are their averages. Also, models are deliberately trained with no augmentation, including rotation as well as typical augmentation methods such as flips or random crop, in order to demonstrate the strict measure of rotation invariance.

\subsection{Metrics}
\vspace{-0.2cm}
\paragraph{Invariance of Representations} Besides the classification accuracy, metrics are defined to evaluate the degree of invariance of the representations. Let $\textbf{z}^{\theta}$ be the representation of the input image $\mathcal{R}^{\theta}$. Then, relative $L_{2}$ norm of rotational variance ($\delta^{\theta}_{L_{2}}$) and absolute cosine similarity of rotational invariance ($\delta^{\theta}_{\cos}$) are defined correspondingly as

\vspace{-0.2cm}
\begin{equation}
    \delta^{\theta}_{L_{2}} = \frac{||\mathbf{z}^{\theta}-\mathbf{z}^{0}||}{||\mathbf{z}^{0}||}, \ \delta^{\theta}_{\cos} = \frac{||\mathbf{z}^{\theta} \cdot \mathbf{z}^{0}||}{||\mathbf{z}^{\theta}|| \ ||\mathbf{z}^{0}||}
\end{equation}

\vspace{-0.1cm}
where $|| \cdot ||$ is the $L_{2}$ norm. 

\begin{figure*}
  \includegraphics[width=\linewidth]{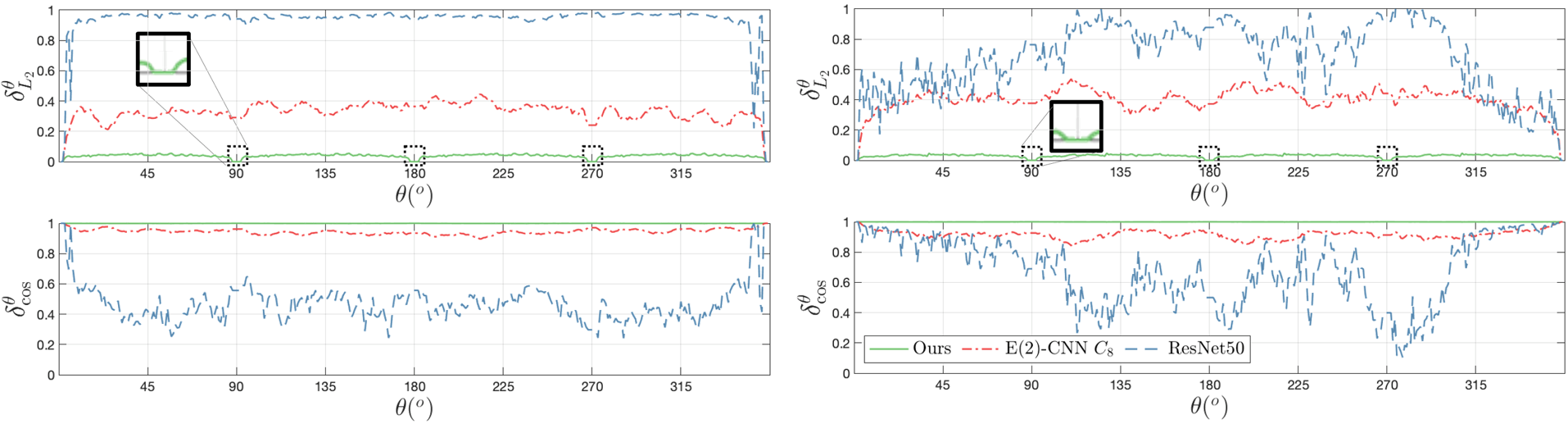}
  \centering

  \vspace{-0.3cm}
  \caption{(L-R): Plots of the averages of rotational $L_{2}$ norm of variance ($\delta^{\theta}_{L_{2}}$) and rotational cosine distance of invariance ($\delta^{\theta}_{\cos}$) of invariant representations over rotation angle $\theta$ for 10 randomly selected images from (L) MNIST and (R) CIFAR-10 dataset. For $\delta^{\theta}_{L_{2}}$, the lower, the better and for $\delta^{\theta}_{\cos}$, the higher, the better.}
  \label{fig:inv}
\end{figure*}

\vspace{-0.2cm}
\section{Results and Discussions} 

\vspace{-0.2cm}
\subsection{Validations over Invariance}
\label{sec:inv}
\vspace{-0.1cm}
\paragraph{Quantitative Validation on Invariance} Given 10 randomly sampled images from MNIST and CIFAR-10 each, $\delta^{\theta}_{L_2}$ and $\delta^{\theta}_{\cos}$ are measured for $\theta \in [0, 360)$ with ResNet-50, E(2)-CNN $C_{8}$, and SWN-GCN, and the averages of the results are reported in Figure \ref{fig:inv}. SWN-GCN produces $\delta^{\theta}_{L_2}$ closer to 0 and $\delta^{\theta}_{\cos}$ closer to 1 over all angles of rotations than other baseline models, which shows that SWN-GCN yields the most rotation-invariant representations out of the baselines. Especially, note that ours yields the exact value of $\delta^{\theta}_{L_2} = 0$ for $\theta$ with multiples of $90^{\circ}$. E(2)-CNN yields significantly better invariance than ResNet-50, but still yields noticeable variances in both $\delta^{\theta}_{L_2}$ and $\delta^{\theta}_{\cos}$ over all rotation angles. 


\vspace{-0.4cm}
\subsection{Rotation-Invariant Image Classification} 
\vspace{-0.2cm}
We also demonstrate the classification accuracy over rotation augmented dataset of our network with other baselines, where all the models are trained with upright images only. As shown in Table \ref{tab:cls_acc}, classification accuracy of our proposed model outperformed all baseline models on the overall accuracy (OA) on test datasets that are augmented with random rotation angles. 



On top of the highest classification accuracy on the rotation augmented dataset, SWN-GCN shows significant improvements in classification accuracy over the largest range of rotation. Table \ref{tab:cls_acc} reports classification accuracies over dataset augmented with fixed angles with multiples of $30^{\circ}$. Even though ResNet-50 shows the highest performance over upright image classification, all other baselines yield higher classification accuracy on rotated images than ResNet-50 in most rotation angles. In particular, SWN-GCN outperforms TIGraNet, the state-of-the-art graph-based isometry-invariant network, in all rotation angles. 

Meanwhile,
one may observe from Table \ref{tab:cls_acc} and Figure \ref{fig:inv} that for some rotation angles, E(2)-CNN yields better $\delta^{\theta}_{L_2}$ and $\delta^{\theta}_{\cos}$ over SWN-GCN yet shows higher classification performances than SWN-GCN (i.e. $\theta= 0^{\circ}, 30^{\circ}, 60^{\circ}$ for MNIST and $\theta = 0^{\circ}, 30^{\circ}$ for CIFAR-10). For these angles, the impact of difference in expressibility of representations between E(2)-CNN and SWN-GCN is more significant than the disruptions on representations introduced by input image rotation on E(2)-CNN. However, disruptions introduced by a moderate amount of input image rotation on E(2)-CNN easily overwhelm the advantage of E(2)-CNN on expressibility, as proven by SWN-GCN outperforming E(2)-CNN in most range of rotations for image classifications.

\begin{table*}[t]
  \caption{Classification accuracies over test datasets rotated by fixed angles. Overall Accuracy (OA) is the result over all range of rotation.}
  \vspace{-0.3cm}
  \label{tab:cls_acc}
  \vskip 0.15in
  \begin{center}
  \begin{small}
  \begin{sc}
    \resizebox{\textwidth}{!}{
    \begin{tabular}{cc|cccccccccccc|c}
    \hline
     & & \multicolumn{13}{c}{Classification Accuracy (\%)}                   \\
    Dataset & Models & $0^{\circ}$ & $30^{\circ}$ & $60^{\circ}$ & $90^{\circ}$ & $120^{\circ}$ & $150^{\circ}$ & $180^{\circ}$ & $210^{\circ}$ & $240^{\circ}$ & $270^{\circ}$ & $300^{\circ}$ & $330^{\circ}$ & OA \\
    
    \hline\hline                                                                               
    MNIST &ResNet-50 \cite{he2016deep}         &  \textbf{99.5}  & 91.9 & 47.7 & 28.1 & 29.7 & 48.8 & 57.4 & 51.1 & 33.7 & 23.5 & 35.2 & 90.0 & 42.4 \\
    
    & E(2)-CNN $C_{8}$ \cite{weiler2019general}                                                                    & 99.3  & \textbf{98.1} & \textbf{95.9} & 96.3 & 86.2 & 74.9 & 70.7 & 71.1 & 81.8 & 95.1 & \textbf{92.9} &\textbf{97.0} & 87.5\\
    
    & TIGraNet \cite{khasanova2017graph} & 89.1  & 82.7 & 79.8 & 89.1 & 82.7 & 79.8  & 89.1 & 82.7 & 79.8 & 89.1 & 82.7 & 79.8 & 85.1 \\ 
    & SWN-GCN                                                                               & 96.5  & 89.8 & 87.3 & \textbf{96.5} & \textbf{89.8} & \textbf{87.3} & \textbf{96.5} & \textbf{89.8} & \textbf{87.3} & \textbf{96.5} & 89.8 & 87.3 & \textbf{91.8} \\

    \hline                                    
    CIFAR-10 &ResNet-50          & \textbf{85.1}  & 54.5 & 34.1 & 18.3 & 27.5 & 26.9 & 35.6 & 27.0 & 24.9 & 33.8 & 33.2 & 52.5 & 36.1  \\
    
    & E(2)-CNN $C_{8}$  & 77.1  & \textbf{57.8} & 44.3 & 48.5 & 34.4 & 30.8 & 37.8 & 31.9 & 35.4 & 49.4 & 45.0 & \textbf{56.0} & 46.2\\
    
    & TIGraNet  & 38.9  & 37.0 & 36.8 & 38.9 & 37.0 & 36.8  & 38.9 & 37.0 & 36.8 & 38.9 & 37.0 & 36.8 & 38.1 \\ 
    & SWN-GCN                                    & 51.3  & 49.6 & \textbf{50.1} & \textbf{51.3} & \textbf{49.6} & \textbf{50.1} & \textbf{51.3}  & \textbf{49.6} & \textbf{50.1} & \textbf{51.3}  & \textbf{49.6} & 50.1 & \textbf{50.5} \\
    \hline
  \end{tabular}}
  \end{sc} 
  \end{small}
  \end{center}
\end{table*} 

Such trade-off of advantages between the two models can be explained by making an analogous comparison with CNN by referring to Eq. \eqref{eq:approx1} for SWN-GCN. All parameters in convolution kernels, i.e. $3 \times 3$, are trainable in CNN, meaning that these kernels are as expressive as they can be. However, these parameters have different values within the convolution kernel and break the local rotation consistency. Whereas for SWN-GCN, only the self-vertex is multiplied with trainable parameters, and the magnitude of adjacency with adjacent vertices are uniform and fixed for the sake of preserving local rotational consistency.

\section{Conclusion}
\vspace{-0.2cm}
We proposed a network that yields equivariant representation with SWN-GCN and invariant representation using GAP. We showed structural equivariance of SWN-GCN and invariance of GAP, and validated the properties with experimental results. Our method achieved the state-of-the-art performances on rotated MNIST and CIFAR-10 image classification, where the models were trained with upright images only.

\bibliography{egbib}
\end{document}